\documentclass[10pt,twocolumn,letterpaper]{article}

\usepackage{iccv,times,graphicx,tabulary,multirow,subfig,xspace,xcolor}
\usepackage{amsthm,amssymb,fixmath,mathtools,nicefrac}
\usepackage[pagebackref=true,breaklinks=true,letterpaper=true,colorlinks,
  citecolor=citecolor,bookmarks=false]{hyperref}
\definecolor{citecolor}{RGB}{34,139,34}

\begin{document}
\title{Towards Good Practices for Multi-Person Pose Estimation}
\track{COCO Keypoint Detection}

\author{
Dongdong Yu$^{\dag}$, Kai Su$^{\dag}$, Changhu Wang \\
ByteDance AI Lab, China \\
}

\maketitle
\thispagestyle{empty}

\let\thefootnote\relax\footnotetext{$^{\dag}$ Equal contribution.}

\begin{abstract}
Multi-Person Pose Estimation is an interesting yet challenging task in computer vision. In
this paper, we conduct a series of refinements with the MSPN and PoseFix Networks, and empirically evaluate their impact on the final model performance through ablation studies. By taking all the refinements, we achieve 78.7 on the COCO test-dev dataset and 76.3 on the COCO test-challenge dataset.
\end{abstract}

\section{Introduction}
Multi-Person Pose Estimation is a fundamental yet challenging problem in computer vision. The goal is to locate body parts for all persons in an image, such as keypoints on the arms, torsos, and the face. It is important for many applications like human re-identification, human-computer interaction and activity recognition.

The tremendous development of deep convolution neural networks \cite{he2016deep} bring huge progress for multi-person pose estimation. Existing approaches can be roughly classified into two frameworks, i.e., top-down framework \cite{li2019rethinking, su2019multi, moon2019posefix, yu2018multi} and bottom-up framework \cite{newell2017associative}. The former one first detects all human bounding boxes in the image and then estimates the pose within each box independently. 
For example, Multi-Stage Pose estimation Network (MSPN) \cite{li2019rethinking} adopts the ResNet-50 through multi stages based on repeated down and up sampling steps. PoseFix Network \cite{moon2019posefix} is a human pose refinement network that refines a estimated pose from a tuple of an input image and a pose. The latter one first detects all body keypoints independently and then assembles the detected body joints to form multiple human poses. For example, Associate Embedding \cite{newell2017associative} designs the network to simultaneously estimate the keypoints detection and group heatmaps, instead of the multi-stage pipelines.

In this paper, we follow the top-down pipeline and conduct a series of refinements based on MSPN and PoseFix Networks and evaluate their impact on the final model performance through ablation studies. Finally, we achieve 78.7 on the COCO test-dev dataset and 76.3 on the COCO test-challenge dataset.

\section{Method}
To handle the multi-person pose estimation, we follow the top-down pipeline. First, a human detector is applied to generate all human bounding boxes in the image. Then we apply pose estimation network to estimate the corresponding human pose.

The MSPN network adopts the ResNet-50 as the backbone of the encoder and decoder. In our work, we propose a new backbone, named Refine-50, which can well handle the scale variant cases.

\section{Experiments}

\subsection{Datasets}

The training datasets include the COCO train2017 dataset \cite{lin2014microsoft} (includes 57K images and 150K person instances) and all the AI Challenge dataset \cite{ai.challenge} (includes 400K person instances). For the AI Challenge dataset, we only use the same annotated keypoints as the COCO train2017 dataset for the training. The final results are reported on the COCO test-dev dataset and the COCO test-challenge dataset.

\subsection{Results}
\subsubsection{Ablation Study}
In this subsection, we will step-wise decompose our model to reveal the effect of each component. In the following experiments, we evaluate all comparisons on the COCO val2017 dataset. We use 4x stage for both MSPN network and our network.

\noindent{\textbf{Effect of Backbone}}~~Different with MSPN, we replace the ResNet-50 with our Refine-50. As shown in Table~\ref{table:table1},  we do the experiment with official MSPN code, the AP is 74.7, which is obvious lower than the MSPN's result from their paper. After replacing the ResNet-50 with our Refine-50, the AP is improved from 74.7 to 76.0. Detection box is provided from the MSPN paper. 

\begin{table}[tb]
	\centering
	\caption{Results with different backbones on COCO val2017 dataset.}\label{table:table1}
	
	\resizebox{1.0\columnwidth}{!}{
	\begin{tabular}{ccc}
		\hline
		Backbone & GT Box & Detection Box \\
		\hline
		ResNet-50(Result from Paper) & 76.5 & 75.9 \\
		ResNet-50(Our implement from github) & 76.2 & 74.7 \\
		Refine-50(Ours) & 77.7 & 76.0 \\
		\hline
	\end{tabular}
	}
\end{table}

\noindent{\textbf{Effect of Image Resolution}}~~By using a larger resolution of input image, the AP performance is improved from 76.0 to 77.5, as shown in Table~\ref{table:table2}. 

\begin{table}[tb]
	\centering
	\caption{Results with different resolution on COCO val2017 dataset.}\label{table:table2}
	
	\resizebox{0.65\columnwidth}{!}{
	\begin{tabular}{ccc}
		\hline
		Resolution & GT Box & Detection Box \\
		\hline
		256x192 & 77.7 & 76.0 \\
		384x288 & 78.9 & 77.5 \\
		\hline
	\end{tabular}
	}
\end{table}

\noindent{\textbf{Effect of Extra Dataset}}~~Besides, we also use extra dataset(AI Challenge Dataset) for training. As shown in Table~\ref{table:table3}, the AP is improved from 77.5 to 78.5 by using the extra dataset.

\begin{table}[tb]
	\centering
	\caption{Results with different training datasets on COCO val2017 dataset.}\label{table:table3}
	
	\resizebox{0.7\columnwidth}{!}{
	\begin{tabular}{ccc}
		\hline
		Training Dataset & GT Box & Detection Box \\
		\hline
		COCO train2017 & 78.9 & 77.5 \\
		COCO and AI & 80.2 & 78.5 \\
		\hline
	\end{tabular}
	}
\end{table}

\subsubsection{Development and Challenge Results}
In this subsection, we ensemble three Refine-50 models for the pose estimation. As shown in Table~\ref{table:table4}, the AP of test-dev is 78.0, and the AP of test-cha is 76.0. After using PoseFix, the AP of test-dev can be improved to 78.7 and the AP of test-cha can be improved to 76.3.

\begin{table}[tb]
	\centering
	\caption{Results of our model on COCO2017 test-dev and test-cha dataset.}\label{table:table4}
	
	\resizebox{0.9\columnwidth}{!}{
	\begin{tabular}{ccc}
		\hline
		BackBone & Development set & Challenge set \\
		\hline
		Refine-50 & 78.0 & 76.0 \\
		Refine-50+PoseFix & 78.7 & 76.3 \\
		\hline
	\end{tabular}
	}
\end{table}

\section{Conclusion}
In this work, we conduct a series of refinements with the MSPN and PoseFix Networks, and empirically evaluate their impact on the final model performance through ablation studies. By taking all the refinements, we achieve 78.7 on the COCO test-dev dataset and 76.3 on the COCO test-challenge dataset.

{\small \bibliographystyle{ieee_fullname} \bibliography{egbib}}

\begin{thebibliography}{1}\itemsep=-1pt

\bibitem{ai.challenge}
AI-Challenge.
\newblock Ai challenge keypoints.
\newblock \url{https://challenger.ai/competition/keypoint/subject}.

\bibitem{he2016deep}
Kaiming He, Xiangyu Zhang, Shaoqing Ren, and Jian Sun.
\newblock Deep residual learning for image recognition.
\newblock In {\em Proceedings of the IEEE conference on computer vision and
  pattern recognition}, pages 770--778, 2016.

\bibitem{li2019rethinking}
Wenbo Li, Zhicheng Wang, Binyi Yin, Qixiang Peng, Yuming Du, Tianzi Xiao, Gang
  Yu, Hongtao Lu, Yichen Wei, and Jian Sun.
\newblock Rethinking on multi-stage networks for human pose estimation.
\newblock {\em arXiv preprint arXiv:1901.00148}, 2019.

\bibitem{lin2014microsoft}
Tsung-Yi Lin, Michael Maire, Serge Belongie, James Hays, Pietro Perona, Deva
  Ramanan, Piotr Doll{\'a}r, and C~Lawrence Zitnick.
\newblock Microsoft coco: Common objects in context.
\newblock In {\em European conference on computer vision}, pages 740--755.
  Springer, 2014.

\bibitem{moon2019posefix}
Gyeongsik Moon, Ju~Yong Chang, and Kyoung~Mu Lee.
\newblock Posefix: Model-agnostic general human pose refinement network.
\newblock In {\em Proceedings of the IEEE Conference on Computer Vision and
  Pattern Recognition}, pages 7773--7781, 2019.

\bibitem{newell2017associative}
Alejandro Newell, Zhiao Huang, and Jia Deng.
\newblock Associative embedding: End-to-end learning for joint detection and
  grouping.
\newblock In {\em Advances in Neural Information Processing Systems}, pages
  2274--2284, 2017.

\bibitem{su2019multi}
Kai Su, Dongdong Yu, Zhenqi Xu, Xin Geng, and Changhu Wang.
\newblock Multi-person pose estimation with enhanced channel-wise and spatial
  information.
\newblock In {\em Proceedings of the IEEE Conference on Computer Vision and
  Pattern Recognition}, pages 5674--5682, 2019.

\bibitem{yu2018multi}
Dongdong Yu, Kai Su, Jia Sun, and Changhu Wang.
\newblock Multi-person pose estimation for pose tracking with enhanced cascaded
  pyramid network.
\newblock In {\em Proceedings of the European Conference on Computer Vision
  (ECCV)}, pages 0--0, 2018.

\end{thebibliography}

\end{document}